\documentclass[conference]{IEEEtran}
\IEEEoverridecommandlockouts
\usepackage{cite}
\usepackage{amsmath,amssymb,amsfonts}
\usepackage{algorithmic}
\usepackage{graphicx}
\usepackage{textcomp}
\usepackage{xcolor}

\usepackage{tabularx}
\usepackage{multirow}
\usepackage{bbding}
\usepackage{pifont}
\usepackage{subcaption}
\usepackage{threeparttable}
\usepackage{booktabs}
\usepackage{makecell}

\definecolor{mygreen}{RGB}{0, 150, 0}
\definecolor{myred}{RGB}{200, 0, 0}

\def\BibTeX{{\rm B\kern-.05em{\sc i\kern-.025em b}\kern-.08em
    T\kern-.1667em\lower.7ex\hbox{E}\kern-.125emX}}
\begin{document}

\title{Research-Oriented Human-Centric Evaluation \\ for Foundation Models }
\author{Yijin Guo$^{1,2}$,  Kaiyuan Ji$^{2,3}$, Xiaorong Zhu$^{1,2}$, Junying Wang$^{2,4}$, \\
Farong Wen$^{1,2}$, Chunyi Li$^{2}$, Zicheng Zhang$^{\dagger2}$, Guangtao Zhai$^{\dagger1,2}$ \\
\\
\textit{$^1$Shanghai Jiao Tong University, $^2$Shanghai AI Laboratory,} \textit{$^3$East China Normal University, $^4$Fudan University}
}


\maketitle

\begin{figure*}[t!]
  \centering
  \includegraphics[width=0.85\textwidth]{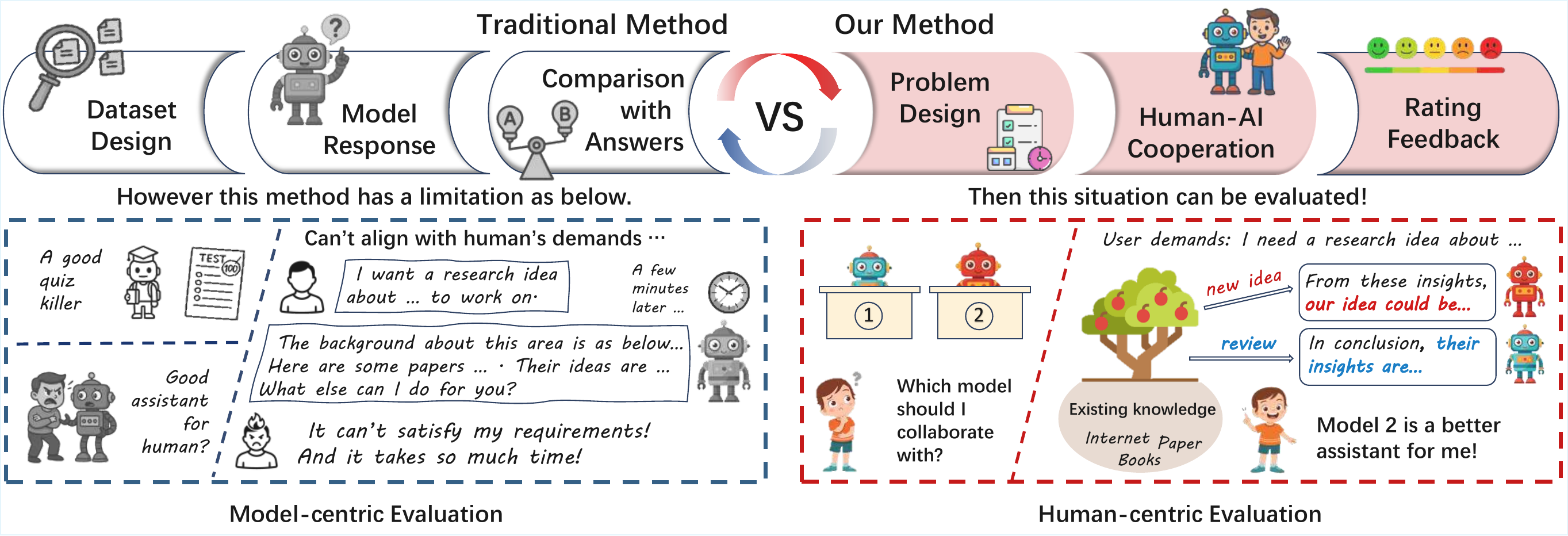} 
  \caption{Motivation for the HCE Framework. Traditional model-centric evaluations primarily focus on quiz-style performance metrics, which do not adequately reflect the human experience. The proposed human-centric evaluation framework better aligns the evaluation of foundation models with the quality of real human interactions.}
  \label{fig:spotlight}
  \vspace{-0.5cm}
\end{figure*}

\begin{abstract}
Most current evaluations of foundation models focus on objective benchmarks, such as knowledge coverage and reasoning accuracy, often overlooking users' subjective experiences in human-AI collaboration. To address this gap, we propose a research-oriented Human-Centric Evaluation framework. It captures user perceptions across three core dimensions: problem-solving ability, information quality, and interaction experience, providing a structured, fine-grained approach to understanding how users evaluate and respond to model behavior in multi-modal research contexts. We conduct 604 human evaluation sessions across various disciplines, involving recent advanced foundation models. Through open-ended, time-limited collaborative tasks, we gather rich subjective assessments that highlight model capabilities and user preferences. Additionally, we perform an LLM-as-a-judge experiment and find that even sophisticated models struggle to accurately replicate human subjective judgment, emphasizing the irreplaceable value of first-person human assessment.
\end{abstract}


\section{Introduction}
The rapid development of artificial intelligence has driven significant progress in large-scale foundation models, which are increasingly expanding from general-purpose applications to various research domains. Current mainstream evaluation approaches primarily rely on \textbf{objective benchmarks}, which assess models through standardized datasets and tasks in terms of knowledge coverage, reasoning ability, or computational efficiency \cite{mmlu, bigbench, gsm, hendrycks2021measuring, chen2021humaneval, swebench, aibench, zhang2025lmmsurvey}. However, such metrics often overlook the \textit{subjective human experience} in real interactions, including ease of use, interaction fluency, and the extent to which models effectively support practical problem solving.  

With the growing adoption of foundation models in real-world scenarios, \textbf{subjective evaluation} has gradually gained attention. Nevertheless, existing efforts often suffer from limited human involvement and coarse-grained evaluation criteria. Some studies have attempted to incorporate perspectives of human values and perception \cite{raza2025humanibench, li2024herm, liao2025humanbeauty, kiela2020hateful, sap2019socialiqa}, but they still rely heavily on predefined benchmarks and lack genuine human participation. In parallel, works such as Chatbot Arena \cite{chatbotarena} and OpenAssistant Conversations \cite{kopf2023openassistant} have introduced real user feedback through arenas and crowdsourcing, offering evaluations closer to authentic interactive experiences. While these explorations provide valuable insights, they often lack multidimensional and fine-grained evaluation designs.  

\begin{table}[t]
    \caption{In comparison to previous work, the proposed HCE framework comprehensively addresses multiple evaluation dimensions, multimodal capabilities, a range of disciplines, and broad application coverage.}
    \label{tab:related_work}
    \centering
    \fontsize{9}{10.5}\selectfont 
    \begin{tabularx}{0.95\linewidth}{@{} l *{2}{c} *{7}{>{\centering\arraybackslash}X} @{}}
        \toprule
        \scalebox{0.85}{\textbf{Method}} & 
        \scalebox{0.85}{\textbf{Time}} & 
        \scalebox{0.85}{\textbf{Type}} &
        \scalebox{0.8}{\textbf{\makecell{M.\\M.}}} &
        \scalebox{0.8}{\textbf{\makecell{A.\\C.}}} &
        \scalebox{0.8}{\textbf{\makecell{PS.\\A.}}} &
        \scalebox{0.8}{\textbf{\makecell{Info.\\Qua.}}} &
        \scalebox{0.8}{\textbf{\makecell{Inter.\\Exp.}}} &
        \scalebox{0.8}{\textbf{\makecell{Hum.}}} &
        \scalebox{0.8}{\textbf{\makecell{Sci.}}} \\
        \midrule
        \scalebox{0.9}{MMLU} & \scalebox{0.9}{20/09} & \scalebox{0.9}{Obj.} & \textcolor{myred}{\ding{55}} & \textcolor{myred}{\ding{55}} & \textcolor{mygreen}{\ding{52}} & \textcolor{myred}{\ding{55}} & \textcolor{myred}{\ding{55}} & \textcolor{mygreen}{\ding{52}} & \textcolor{mygreen}{\ding{52}} \\
        
        \scalebox{0.9}{SWE-bench} & \scalebox{0.9}{23/10} & \scalebox{0.9}{Obj.} & \textcolor{myred}{\ding{55}} & \textcolor{myred}{\ding{55}} & \textcolor{mygreen}{\ding{52}} & \textcolor{mygreen}{\ding{52}} & \textcolor{myred}{\ding{55}} & \textcolor{mygreen}{\ding{52}} & \textcolor{myred}{\ding{55}} \\
        
        \scalebox{0.9}{GPQA\cite{gpqa}} & \scalebox{0.9}{23/11} & \scalebox{0.9}{Obj.} & \textcolor{myred}{\ding{55}} & \textcolor{myred}{\ding{55}} & \textcolor{mygreen}{\ding{52}} & \textcolor{myred}{\ding{55}} & \textcolor{myred}{\ding{55}} & \textcolor{mygreen}{\ding{52}} & \textcolor{myred}{\ding{55}}\\
        
        \scalebox{0.85}{Chatbot Arena} & \scalebox{0.9}{24/03} & \scalebox{0.9}{Arena} & \textcolor{myred}{\ding{55}} & \textcolor{mygreen}{\ding{52}} & \textcolor{mygreen}{\ding{52}} & \textcolor{mygreen}{\ding{52}} & \textcolor{mygreen}{\ding{52}} & \textcolor{myred}{\ding{55}} & \textcolor{myred}{\ding{55}}  \\
        
        \scalebox{0.9}{HLE\cite{hle}} & \scalebox{0.9}{25/01} & \scalebox{0.9}{Obj.} & \textcolor{mygreen}{\ding{52}} & \textcolor{myred}{\ding{55}} & \textcolor{mygreen}{\ding{52}} & \textcolor{myred}{\ding{55}} & \textcolor{myred}{\ding{55}} & \textcolor{mygreen}{\ding{52}} & \textcolor{mygreen}{\ding{52}}  \\
        
        \scalebox{0.9}{Our method} & -- & \scalebox{0.9}{Subj.} & \textcolor{mygreen}{\ding{52}} & \textcolor{mygreen}{\ding{52}} & \textcolor{mygreen}{\ding{52}} & \textcolor{mygreen}{\ding{52}} & \textcolor{mygreen}{\ding{52}} & \textcolor{mygreen}{\ding{52}} & \textcolor{mygreen}{\ding{52}}  \\
        \bottomrule
    \end{tabularx}
    \begin{tablenotes}
        \footnotesize
        \item M.M. = Multi-Modal; A. C. = Application Coverage; PS.A. = Problem-Solving Ability; Info. Qua. = Information Quality; Inter. Exp. = Interaction Experience; Hum. = Humanities; Sci. = Sciences.
    \end{tablenotes}
    \vspace{-0.5cm}
    
\end{table}

To address the gap in evaluating foundation models, we propose the \textbf{Human-Centric Evaluation} (HCE) framework, focusing on \textit{problem-solving ability}, \textit{information quality}, and \textit{interaction experience}. Designed for research-oriented tasks such as literature review and hypothesis generation, this framework emphasizes both strong knowledge and effective collaboration. We conduct 604 human evaluations through open-ended, time-limited human-model collaborative tasks, comparing multiple advanced foundation models. Our analysis, including pearson correlation coefficients and confirmatory factor analyses, validates the multidimensional structure of the framework, demonstrating that model performance depends on balanced capabilities across dimensions, with rankings consistent across different disciplines.

Additionally, we perform an \textbf{LLM-as-a-judge} experiment, where several advanced language models are tasked with replicating human ratings using the same evaluation criteria. The results show that even the most advanced models struggle to accurately simulate human subjective judgment, emphasizing the ongoing necessity of human evaluation for capturing nuanced, process-oriented aspects of model performance.


Our main contributions can be summarized as follows:  
\begin{itemize}  
    \item We propose and validate a human-centric evaluation framework for foundation models, which enables a systematic and comprehensive subjective assessment of their performance in research-oriented scenarios.
    \item We construct a dataset of 604 human-labeled evaluations across multiple research domains, laying the groundwork for future automated subjective evaluation methods.
    \item We conduct an LLM-as-a-judge experiment showing that current foundation models fail to replicate human subjective ratings, underscoring the limits of automated evaluation and the ongoing need for human assessment.
\end{itemize}

\vspace{0.5em}
\section{Related Work}

\subsection{Benchmark and Objective Datasets}
Benchmarks for foundation models provide standardized evaluations and are central to driving technical progress. Objective datasets, as represented by MMLU, GSM8K, and others \cite{mmlu, gsm, swebench, bigbench, hendrycks2021measuring, chen2021humaneval}, effectively measure model performance on tasks such as knowledge recall, reasoning, and code generation, thereby promoting model comparability and iterative improvement \cite{bert, llama}. However, these evaluations primarily concentrate on task accomplishment and struggle to adequately reflect users' subjective experience in real interactions.

\vspace{0.5em}
\subsection{Subjective Attempts}

As models are increasingly deployed in practical settings, researchers have begun to explore human-centric evaluation. On one hand, works such as Humanibench, UniAA, and others \cite{raza2025humanibench, zhou2024uniaa, li2024herm, liao2025humanbeauty, kiela2020hateful, sap2019socialiqa, shen2024empathicstories++} attempt to assess the alignment of models with human values across dimensions like ethics and aesthetics, yet they still rely on predefined objective benchmarks. On the other hand, methods like Chatbot Arena \cite{chatbotarena, kopf2023openassistant} directly incorporate human ratings. These approaches are closer to real-world experience and can better capture interaction quality and user perception, but their evaluation dimensions are often coarse-grained and fail to systematically disentangle different aspects of subjective perception. In summary, the scale and influence of these works highlight an urgent need for a unified framework to systematically evaluate the comprehensive assistive capabilities of foundation models in realistic scenarios.
\begin{figure*}[]
  \centering
  \includegraphics[width=0.85\textwidth]{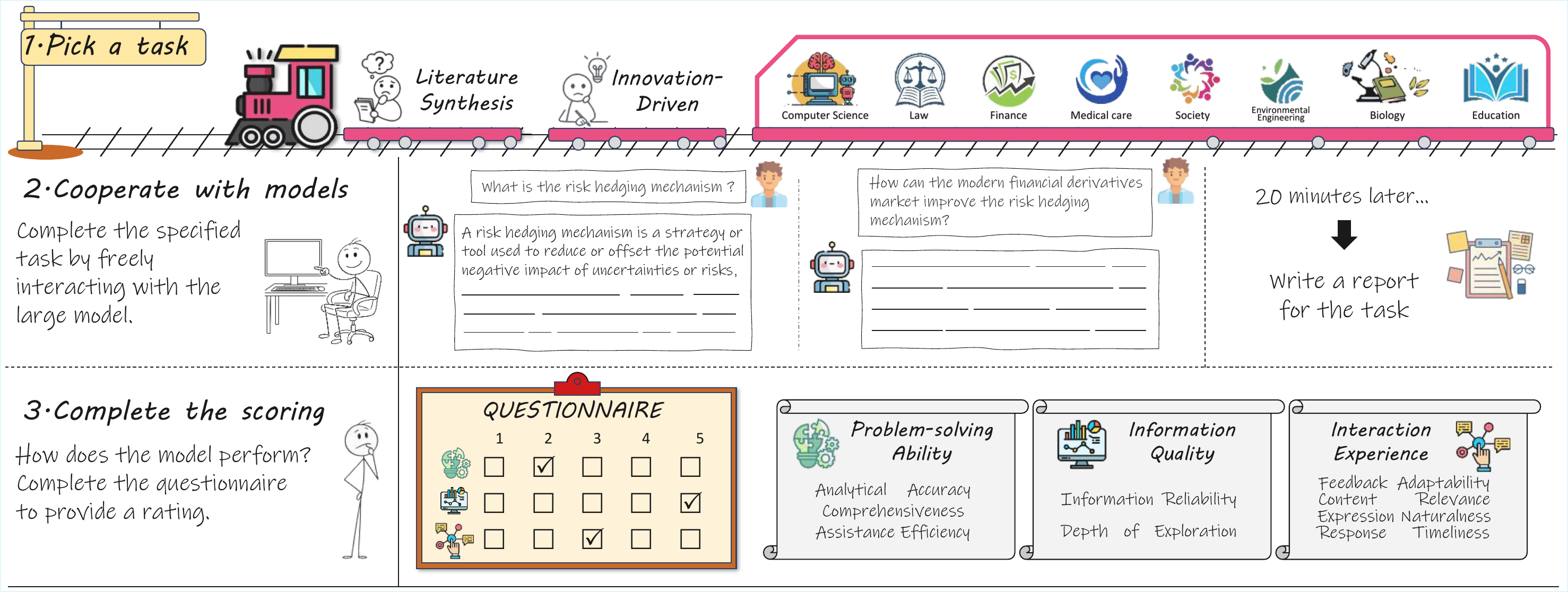} 
  \caption{HCE Experiment Framework. Participants select a task based on their field of study and interests, interact freely with a foundation model for 20 minutes to complete the task, and then complete a questionnaire to evaluate the model's performance.}
  \label{fig:framework}
\end{figure*}

\section{Construction}

\subsection{Design Principles}
To overcome the limitations of traditional objective evaluation in measuring real-world user experience, this study proposes a human-centric subjective evaluation framework. Our core philosophy is that a valid assessment of assistive capability must return to \textit{authentic human-AI collaboration scenarios} and \textit{directly capture the user's subjective experience} during task completion. To this end, our design follows these key principles. First, the evaluation dimensions are derived from user perception to systematically deconstruct different facets that contribute to the interaction experience. Second, tasks must be tailored to the user's background and actual needs to stimulate genuine engagement and interaction. Finally, the tasks themselves should possess sufficient openness and complexity to align with real-world scenarios, moving beyond simple Q\&A. The design of the evaluation dimensions is detailed in \ref{sec:dimension}, followed by the task design in \ref{sec:task}.


\subsection{Evaluation Dimensions}
\label{sec:dimension}

\begin{figure*}[]
  \centering
  \includegraphics[width=0.85\textwidth]{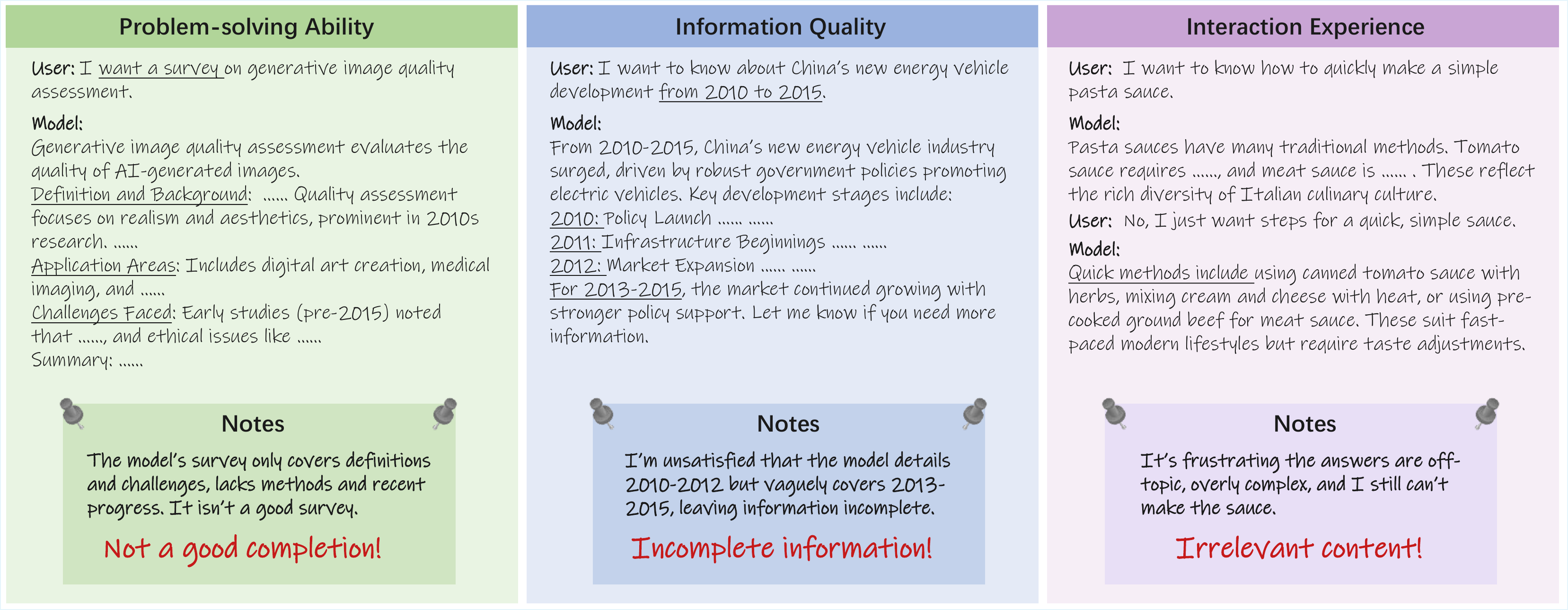} 
  \caption{Examples of Limitations Across HCE Dimensions. This figure presents specific examples of user-model interactions for each evaluation dimension. Accompanying notes highlight key issues in the responses of the foundation models.}
  \label{fig:bad case}
\end{figure*}

Synthesizing the findings of numerous studies, we propose a novel framework for subjective assessment, incorporating three key perspectives: \textit{\textbf{problem-solving capability}}, \textit{\textbf{information accuracy and quality}}, and \textit{\textbf{user interaction experience}. Figure \ref{fig:bad case}} presents some failure cases for these dimensions.

\subsubsection{{Problem-Solving Ability}} This is a critical dimension for evaluating the capacity of a foundation model to analyze, reason, and effectively resolve diverse problems. Building on prior objective evaluations \cite{helm, mrasurvey}, our framework assesses this ability through three dimensions: \textit{\textbf{analytical accuracy}} referring to the precision of the model's reasoning and resolution, \textit{\textbf{comprehensiveness}} measuring the extent to which all relevant problem aspects are considered, and \textit{\textbf{assistance efficiency}} evaluating how effectively the model conserves the user's time and effort in reaching a solution.

\subsubsection{{Information Quality}} It serves as a core dimension for assessing the credibility, completeness, and depth of content generated by foundation models, especially in complex or high-stakes tasks. Grounded in the prior emphasis that high-quality outputs depend on both reliable information and profound knowledge \cite{guo2023evaluating}, our framework assesses this dimension through two constituent aspects: \textit{\textbf{information reliability}} which pertains to the factual accuracy, consistency, and currency of the provided content, and \textit{\textbf{exploration depth}} which captures the level of detail, contextual relevance, and thoroughness in the information and its accompanying analysis.

\subsubsection{{Interaction Experience}} It constitutes a pivotal evaluation criterion for foundation models in human-computer interaction, focusing on their capacity to deliver fluent, relevant, and efficient responses. Taori et al. (2023) underscore that interaction experience is evaluated through instruction-following robustness, task relevance, linguistic fluency, and response latency \cite{alpaca}, while Chan et al. (2024) highlight contextual consistency and dynamic responsiveness as further key determinants \cite{chateval}. Grounded in these studies, our framework defines four subjective evaluation dimensions: \textit{\textbf{content relevance}} refers to how closely the model’s responses align with the user’s query, \textit{\textbf{feedback adaptability}} indicates the model’s ability to adjust responses based on user feedback, \textit{\textbf{expression naturalness}} assesses whether responses are natural, coherent, and highly readable, and \textit{\textbf{response timeliness}} measures how promptly and efficiently the model delivers answers.

\subsection{Task Design}
\label{sec:task}


Research-oriented tasks are chosen for evaluation because they inherently involve a wide range of diverse and multimodal content, such as text, codes, and figures, and require a comprehensive set of skills, including reasoning and knowledge synthesis. This makes them an ideal context for systematically assessing model capabilities across various domains.

Two key research problem types are designed. \textit{\textbf{Literature Synthesis}} Problems simulate the need for literature reviews in research, comprehensively tasking the model's ability to organize and summarize domain knowledge, evaluating its ability of information integration and logical structuring. 
\textit{\textbf{Innovation-Driven}} Problems focus on interdisciplinary innovation, requiring the model to propose novel hypotheses or solutions by identifying research gaps, assessing its creative thinking, knowledge transfer, and deep reasoning skills.

Tasks span eight disciplines: \textit{\textbf{Computer Science, Law, Finance, Medicine, Sociology, Environmental Engineering, Biology, and Education}}. The selection reflects diverse research paradigms, which tests models’ adaptability to varied methodologies and prevents overfitting to single-domain biases, ensuring a model’s robustness as a research assistant.


\begin{table*}[t]
    \caption{Subjective leaderboard across dimensions and disciplines. Best in \textbf{bold}. Sub-dimensions are abbreviated\protect\footnotemark.}
    \label{tab:leaderboard_merged}
    \centering
    \renewcommand\arraystretch{1.1}
    \setlength{\tabcolsep}{5pt}
    \resizebox{\textwidth}{!}{
    \begin{tabular}{
    c|l|c|
    c c c|
    c c|
    c c c c|
    c c c c c c c c
    }
    \hline
    \hline
    \multirow{2}{*}{\textbf{Rank}} &
    \multirow{2}{*}{\textbf{Model}} &
    \multirow{2}{*}{\textbf{Avg.}} &
    \multicolumn{3}{c|}{\textbf{Problem-Solving Ability}} &
    \multicolumn{2}{c|}{\textbf{Information Quality}} &
    \multicolumn{4}{c|}{\textbf{Interaction Experience}} &
    \multicolumn{8}{c}{\textbf{Disciplines}} \\
    \cline{4-6} \cline{7-8} \cline{9-12} \cline{13-20}
     & & &
     A.A. & Compre. & A.E. &
     I.R. & E.D. &
     C.R. & F.A. & E.N. & R.T. &
     AI & Law & Fin. & Med. & Soc. & Env. & Bio. & Edu. \\
    \hline
    1 & Grok-3 & \textbf{4.30}
    & 4.25 & \textbf{4.41} & \textbf{4.13}
    & 4.19 & \textbf{4.26}
    & \textbf{4.43} & \textbf{4.19} & 4.40 & \textbf{4.44}
    & \textbf{4.30} & \textbf{4.41} & 4.10 & \textbf{4.15} & \textbf{4.26} & \textbf{4.41} & \textbf{4.41} & \textbf{4.24} \\

    2 & Gemini-2.5 & 4.09
    & \textbf{4.30} & 4.21 & 3.59
    & \textbf{4.26} & 3.98
    & 4.15 & 3.64 & \textbf{4.44} & 4.22
    & 4.18 & 4.20 & \textbf{4.28} & 3.92 & 4.23 & 4.00 & 4.12 & 3.80 \\
    
    3 & DeepSeek-R1 & 4.08
    & 4.09 & 4.21 & 3.80
    & 4.18 & 4.10
    & 4.15 & 3.75 & 4.26 & 4.17
    & 4.16 & 4.00 & 4.11 & 3.90 & 4.15 & 4.18 & 4.10 & 4.07 \\

    4 & OpenAI o3-mini & 3.91
    & 4.02 & 3.66 & 4.05
    & 4.02 & 3.45
    & 4.12 & 4.09 & 3.88 & 3.90
    & 4.04 & 3.84 & 3.77 & 3.86 & 3.88 & 3.86 & 3.89 & 3.92 \\
    \hline
    \hline
    \end{tabular}
    }
\end{table*}

\section{Subjective Experiments}

\subsection{Setup}

\subsubsection{Procedure} This study employs a \textit{time-limited, free-form Q\&A interactive paradigm} for evaluation. Each evaluator first establishes a dialogue connection with a selected large model. Subsequently, within a {20-minute} time limit, they engage in multiple rounds of free-form Q\&A interaction centered around their task, with the goal of completing the task and finding a solution to advance the task progressively.

\subsubsection{Evaluators and Models} The participants involved in the evaluation are current or graduated postgraduate students with experience in using large models (typically more than three months). They select corresponding test tasks based on their own academic backgrounds and interact with the models throughout using their proficient language (Chinese or English). The experiment evaluates 4 advanced large models: DeepSeek-R1 \cite{deepseek}, OpenAI o3-mini \cite{o3}, Grok-3 \cite{grok}, and Gemini-2.5 \cite{gemini}. All interactions are conducted through official interfaces and support multimodal content.

\subsubsection{Rating Mechanism} The scoring strictly revolves around the evaluation dimensions defined in \ref{sec:dimension}. After completing each dialogue task, evaluators independently score each dimension based on their subjective experience, using discrete 5-point scale (1: Very Weak, 2: Weak, 3: Moderate, 4: Strong, 5: Very Strong). The final score is calculated by MOS \cite{bt500}.

\subsubsection{Experimental Scale and Design} Each model, for each language and eachappend type of task, is tested by 5 evaluators. This study completes a total of 604 valid\footnote{Outliers are removed by excluding raters whose Spearman rank correlation coefficient (SRCC) with other raters for a given model fell below 0.4.} independent dialogue and rating instances. The experimental design follows a comparative principle: the same group of evaluators is required to interact with all different models for the same task and provide independent ratings for each. A total of 82 evaluators participate in the entire experimental process.

\subsection{Analysis \& Results}

\subsubsection{Dimensional Analysis of Framework}
\label{sec:dimensional_analysis}

\begin{figure}[t]
  \centering
  \includegraphics[width=0.78\linewidth]{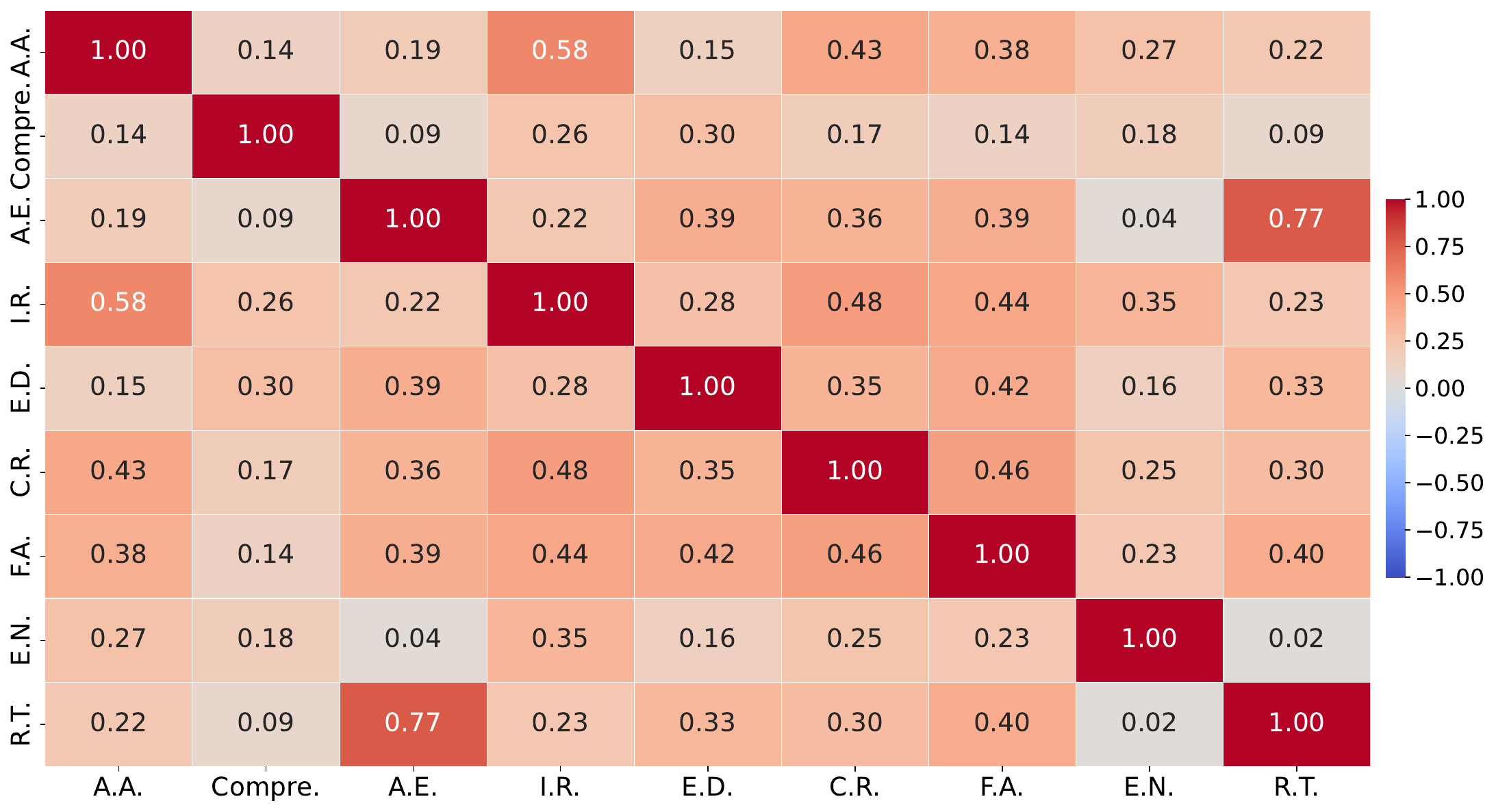} 
  \caption{Pearson correlation heatmap across dimensions.}
    \label{fig:pearson_ds}
\end{figure}

\begin{figure}[t]
  \centering
  \includegraphics[width=0.75\linewidth]{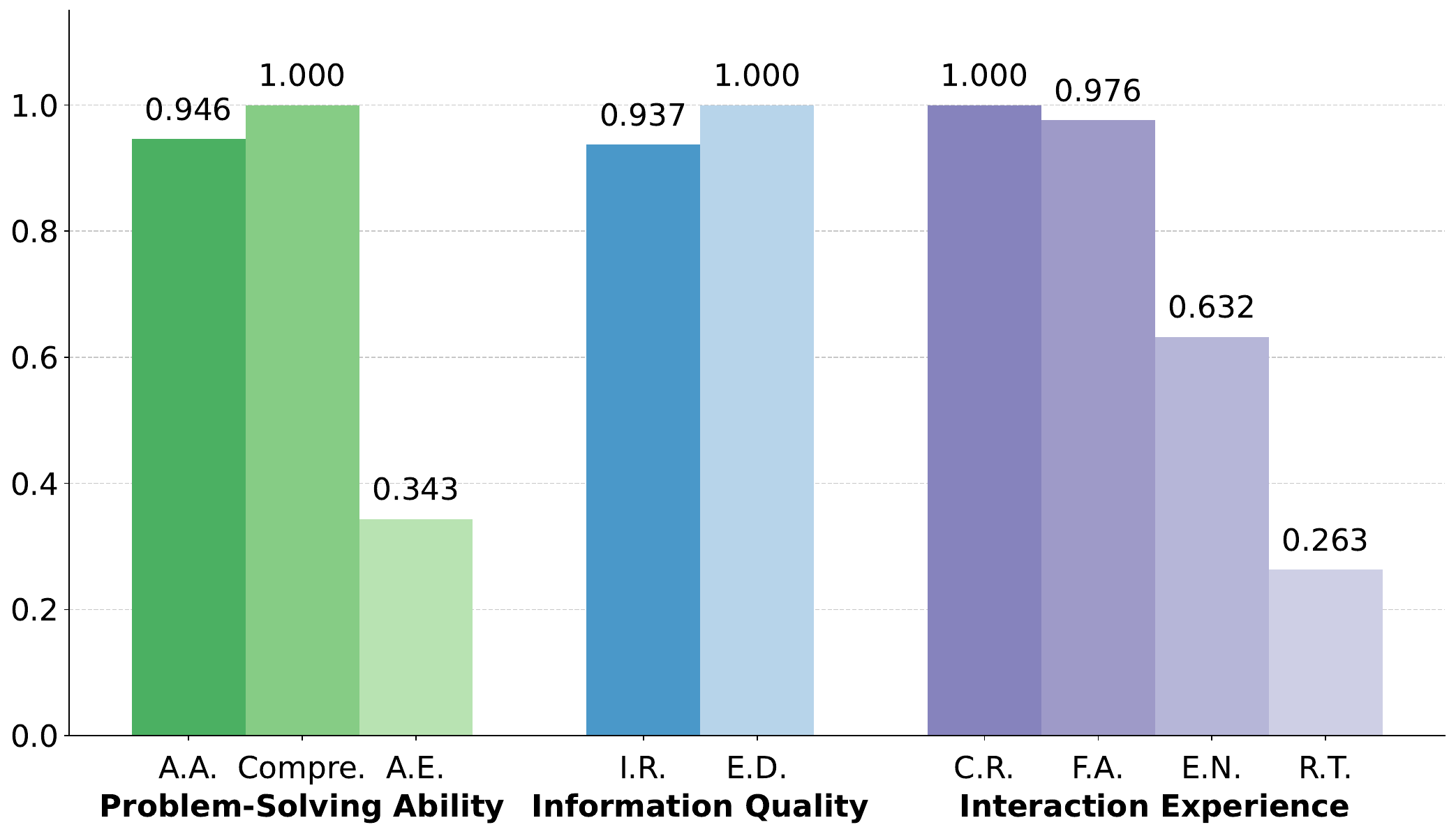} 
  \caption{Standardized factor loadings from factor analysis.}
  \label{fig:factor_loadings}
\end{figure}

To validate the multidimensional framework, several analytical methods are applied. Pearson correlation analysis \cite{Pearson1895} is used to examine the relationships among the nine sub-dimensions in Figure \ref{fig:pearson_ds}, while confirmatory factor analysis \cite{joerskog1969} tests the grouping of these sub-dimensions into three higher-order constructs, with standardized loadings in Fig. \ref{fig:factor_loadings}.

The correlation matrices reveal low to moderate positive correlations (0.1 to 0.4) among sub-dimensions, indicating they capture distinct aspects of model performance, supporting the framework’s discriminative validity. Confirmatory factor analysis shows strong loadings ($\ge$ 0.5) on their assigned higher-order constructs, confirming that the framework decomposes subjective evaluation into coherent, measurable facets.

The two efficiency-related dimensions, assistance efficiency and response timeliness, show a strong correlation but lower loadings on their higher-order constructs. This suggests response timeliness disproportionately influences perceived assistance efficiency, reflecting \textbf{the distinction between outcome-oriented dimensions} (e.g., accuracy, relevance) \textbf{and time-sensitive process characteristics} (e.g., efficiency, timeliness), which are assessed independently in human evaluation, aligning with prior work in NLP and HCI \cite{international2018ergonomics, walker1997paradise}.

\begin{table*}[]
\caption{Accuracy of LLM-as-a-judge evaluations across dimensions. Best in \textbf{bold}.}
\label{tab:llm_as_a_judge_performance}
\renewcommand\arraystretch{1.1}
\setlength{\tabcolsep}{10pt}
\resizebox{\textwidth}{!}{
\begin{tabular}{l|c|ccc|cc|cccc}
\hline \hline
\multicolumn{1}{c|}{\multirow{2}{*}{\textbf{Model}}} & \multirow{2}{*}{\textbf{Avg.}} & \multicolumn{3}{c|}{\textbf{Problem-Solving Ability}}                    & \multicolumn{2}{c|}{\textbf{Information Quality}}    & \multicolumn{4}{c}{\textbf{Interaction Experience}}                                               \\ \cline{3-11} 
\multicolumn{1}{c|}{}                       &                              & \textit{A.A.} & Compre. & A.E. & I.R. & E.D. & C.R. & F.A. & E.N. & R.T. \\ \hline
GPT-5.2                                     & \textbf{0.515}                        & \textbf{0.562}               & \textbf{0.535}             & \textbf{0.535}                 & 0.340                    & \textbf{0.556}             & \textbf{0.521}             & \textbf{0.535}                 & \textbf{0.472}                  & \textbf{0.576}               \\ 
GPT-4o                                      & 0.424                        & 0.361               & 0.438             & 0.382                 & 0.396                   & 0.389             & 0.417             & 0.500                   & 0.354                  & \textbf{0.576}               \\ 
Claude-3-7-Sonnet                  & 0.417                        & 0.426               & 0.333             & 0.351                 & \textbf{0.440}                    & 0.505             & 0.411             & 0.480                  & 0.278                  & 0.533               \\ 
Grok-4-1-Fast-Reasoning                     & 0.380                        & 0.292               & 0.389             & 0.312                 & 0.340                    & 0.354             & 0.431             & 0.382                 & 0.382                  & 0.535               \\ 
DeepSeek-V3.2                               & 0.376                        & 0.340                & 0.403             & 0.347                 & 0.315                   & 0.389             & 0.410              & 0.364                 & 0.368                  & 0.451               \\ 
Qwen3-235B-A22B                             & 0.366                        & 0.302               & 0.388             & 0.272                 & 0.384                   & 0.367             & 0.434             & 0.336                 & 0.399                  & 0.413               \\ 
Gemini-3-Flash-Preview                      & 0.365                        & 0.300                 & 0.366             & 0.343                 & 0.360                    & 0.362             & 0.420              & 0.354                 & 0.345                  & 0.431               \\ 
GPT-OSS-120B                                & 0.338                        & 0.275               & 0.409             & 0.309                 & 0.291                   & 0.346             & 0.411             & 0.326                 & 0.353                  & 0.319               \\ 
Qwen3-8B                               & 0.329                        & 0.277               & 0.354             & 0.262                 & 0.292                   & 0.357             & 0.415             & 0.357                 & 0.357                  & 0.289               \\ 
Gemma-3-27B-IT                       & 0.322                        & 0.261               & 0.363             & 0.265                 & 0.290                    & 0.355             & 0.391             & 0.328                 & 0.362                  & 0.283               \\ \hline \hline
\end{tabular}
}
\end{table*}

\subsubsection{Subjective Experiment Results}
Based on 604 subjective evaluations, sub-dimension scores are averaged and aggregated across evaluation dimensions and disciplines, with the results summarized in Table \ref{tab:leaderboard_merged}. The results show that overall subjective performance is driven more by balanced performance across multiple evaluation dimensions, rather than by exceptional strength in any single capability. Grok-3 ranks highest primarily because it performs consistently well across problem-solving ability, information quality, and interaction experience. By comparison, Gemini-2.5 and DeepSeek-R1 achieve similar overall scores despite different strengths, suggesting that pronounced strengths in some dimensions cannot fully compensate for weaknesses in other evaluation.

Across disciplines, model performance varies in absolute scores, but for the general-purpose foundation models evaluated here, disciplinary differences mainly affect the extent of performance gaps rather than the relative ordering of models. As shown in Table \ref{tab:leaderboard_merged}, models tend to maintain similar rankings across disciplines, while score differences expand or contract depending on the discipline. This suggests that, in subjective human evaluation of general-purpose models, disciplinary context modulates how models’ general reasoning and interaction characteristics are expressed, instead of directly reflecting specialized domain competence.

\footnotetext{A.A. (Analysis Accuracy), Compre. (Comprehensiveness), and A.E. (Assistance Efficiency); I.R. (Information Reliability) and E.D. (Exploration Depth); C.R. (Content Relevance), F.A. (Feedback Adaptability), E.N. (Expression Naturalness), and R.T. (Response Timeliness)}

\section{Benchmark}

\subsection{Setup}
As large language models have been used as judges in automatic evaluation, we conduct an LLM-as-a-judge experiment to examine whether such models can simulate human subjective ratings under the evaluation framework of this study. All human evaluators are trained with the same guidelines and scoring criteria, allowing us to assume that rating standards are aligned across evaluators and that observed score variations primarily reflect differences in interaction content. Under this assumption, multiple LLMs are employed as judges and asked to assign dimension-level scores to complete human–model interaction logs using the same evaluation criteria and a unified prompt, without access to human ratings. 

In this experiment, human ratings are first normalized and divided into three categories: low, medium, and high. Multiple LLMs are employed as judges and tasked with predicting these 3-level ratings for the same interactions. The models used in the experiment include GPT-5.2 \cite{gpt5.2}, DeepSeek-V3.2 \cite{deepseekai2025deepseekv32}, GPT-4o \cite{openai_gpt4o_2024}, Qwen3-8B, Qwen3-235B-A22B \cite{qwen3}, GPT-OSS-120B \cite{agarwal2025gptoss}, Grok-4-1-Fast-Reasoning \cite{grok4_1_xai_2025}, Gemini-3-Flash-Preview \cite{gemini3}, Claude-3-7-Sonnet \cite{anthropic_claude3.7_sonnet_2025}, and Gemma-3-27B-IT \cite{team2025gemma}. Each model is required to assign a 3-level score to the human-model interaction logs, based on the same evaluation criteria and a unified prompt. The accuracy of each model’s predictions is then measured by comparing the predicted scores to the normalized human ratings.

\subsection{Results}

The results in Table \ref{tab:llm_as_a_judge_performance} show that even the most advanced LLMs, such as GPT-5.2 with a total accuracy of 0.515, struggle to replicate human ratings. Many models, including DeepSeek-V3.2 and Qwen3-235B-A22B, show much lower accuracy, with some scores approaching random chance (around 0.333). Overall, the accuracy of all models remains significantly lower than expected for effective human-like evaluation.

The low accuracy suggests that \textbf{automated evaluation models are not yet reliable substitutes for human assessments within the multi-dimensional framework}. This discrepancy can be attributed to the structural limitations of LLMs, which lack interaction experience and temporal awareness. While LLMs rely on static text features and semantic cues, human evaluations often depend on dynamic, process-oriented factors, such as interaction efficiency and overall user experience. Therefore, LLMs are better suited for content-based evaluations, and their applicability in multi-dimensional subjective assessments remains limited. Human evaluation continues to be essential in such complex assessments.

\section{Conclusion}

This study proposes a human-centric subjective evaluation framework to address the limitations of objective benchmarks in assessing foundation models’ real-world usability. By introducing user-perception–driven dimensions and conducting subjective experiments on some advanced foundation models using diverse multimodal tasks, we construct a dataset of 604 human evaluations that reflects authentic user interactions. Our results show that current models fail to reliably approximate human judgments, as evidenced by the LLM-as-a-judge experiments, underscoring the necessity of human-centered evaluation for capturing nuanced model performance. Overall, this framework complements traditional benchmarks by providing a user-oriented perspective to inform model assessment. Future work may explore scalable subjective evaluation through AI-assisted scoring combined with human feedback, as well as tighter integration of subjective and objective metrics.

\newpage
\bibliographystyle{IEEEbib}
\bibliography{content/reference}

\newpage
\section{Appendix}

\subsection{Participant Recruitment and Demands}

Participants are recruited from individuals currently pursuing or recently graduated with a master’s or doctoral degree, with a preference for those who have prior experience using large-scale foundation models, typically for at least three months. To ensure a diverse representation of professional backgrounds, individuals with relevant expertise in research or technical fields are prioritized.

The total compensation for participant involvement is approximately \$2100, covering training and participation in the evaluation tasks. Participants are compensated for their time and effort in accordance with standard labor rates.

Before the experiment, participants undergo a training session to familiarize themselves with the evaluation process and expectations. During this session, they are introduced to the details of the evaluation dimensions. We also provide a sample dialogue for practice, and participants are required to score the dialogue. They are allowed to proceed to the formal experiment only if their scoring aligns closely with the standard answers. Moreover, each participant is required to submit deliverables for each task, which are reviewed for completeness and quality. Only those tasks that meet the predefined criteria are considered valid samples for evaluation.

\subsection{Inter-Rater Agreement Test}

To evaluate the consistency of ratings across different annotators, we adopt the 3$\sigma$ method as a statistical test of inter-rater agreement. This approach, grounded in the theory of normal distribution, assumes that when rating errors follow a Gaussian pattern, approximately 99.73\% of the observations should lie within three standard deviations of the overall mean. Let $\mu=\frac{1}{n}\sum_{i=1}^{n}x_i$ denote the sample mean and $\sigma=\sqrt{\frac{1}{n-1}\sum_{i=1}^{n}(x_i-\mu)^2}$ the standard deviation. Ratings falling outside the interval $[\mu-3\sigma,\;\mu+3\sigma]$ are regarded as potential inconsistencies, while ratings contained within this range indicate satisfactory agreement among raters.

\begin{figure}[htbp]
    \centering
    \includegraphics[width=0.8\linewidth]{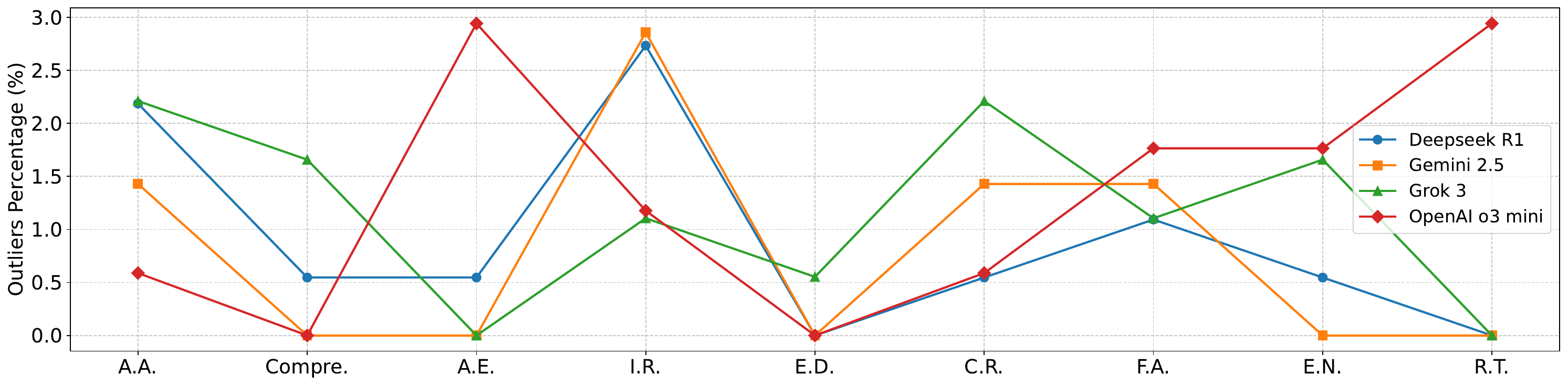}
    \caption{Consistency test results for subjective experimental data across different models.}
    \label{fig:outliers_percentage_plot}
\end{figure}

Figure~\ref{fig:outliers_percentage_plot} presents the proportion of outliers detected by the 3$\sigma$ test across different models and evaluation dimensions. Overall, most dimensions exhibit outlier rates below 3\%, which aligns with the theoretical expectation of the 3$\sigma$ criterion and demonstrates strong overall consistency among human ratings. Notably, dimensions related to \textit{interaction experience} (such as response timeliness and expression naturalness) show slightly higher outlier rates than those of \textit{problem-solving ability} and \textit{information quality}, indicating that annotators perceive greater variability in interactive aspects such as fluency and speed. These results not only validate the robustness of the evaluation protocol but also reveal subtle differences in rating stability across dimensions. Additional detailed results are provided in the supplementary material.

\subsection{More Performance comparison in subjective experiments.}
\label{app:analysis}

\begin{figure}[t]
  \centering
  \begin{subfigure}[t]{0.48\columnwidth}
    \centering
    \includegraphics[width=\linewidth]{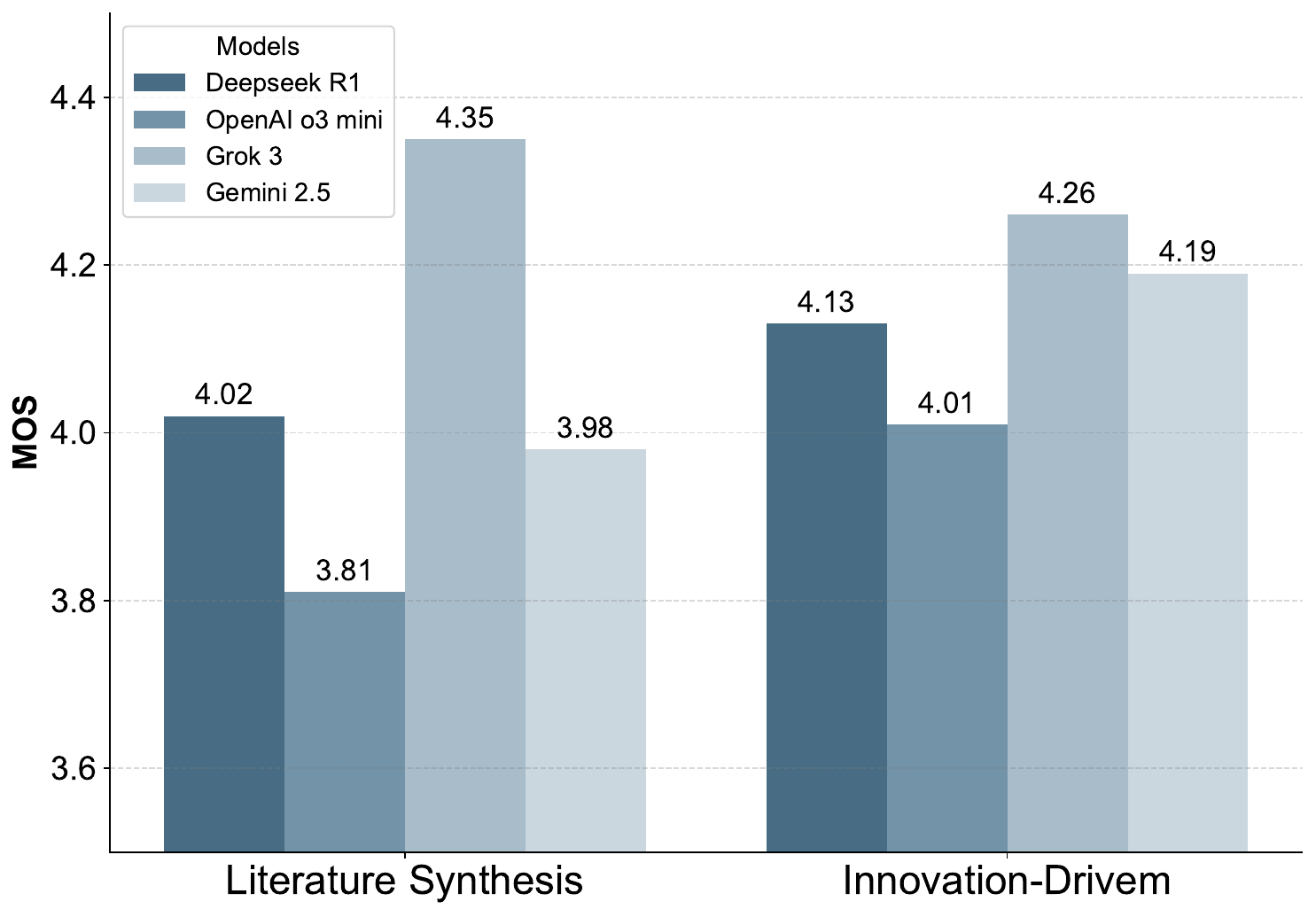}
    \caption{Problem type}
    \label{fig:problem}
  \end{subfigure}%
  \hfill
  \begin{subfigure}[t]{0.48\columnwidth}
    \centering
    \includegraphics[width=\linewidth]{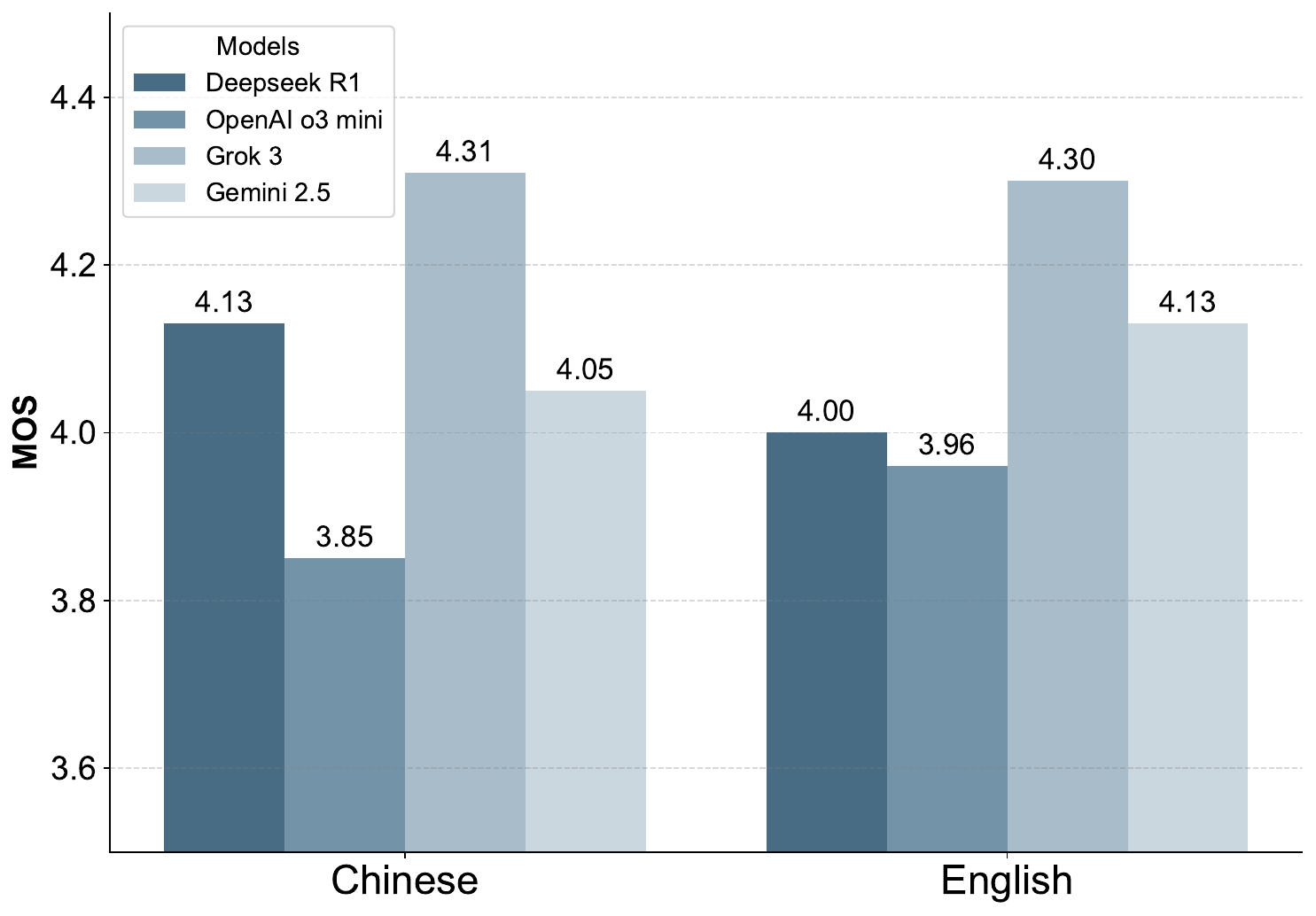}
    \caption{Language}
    \label{fig:language}
  \end{subfigure}
  \caption{Performance for problem types and languages.}
  \label{fig:combined results}
\end{figure}

\begin{figure}
    \centering
    \begin{subfigure}{0.49\columnwidth}
    \centering
    \includegraphics[width=\linewidth]{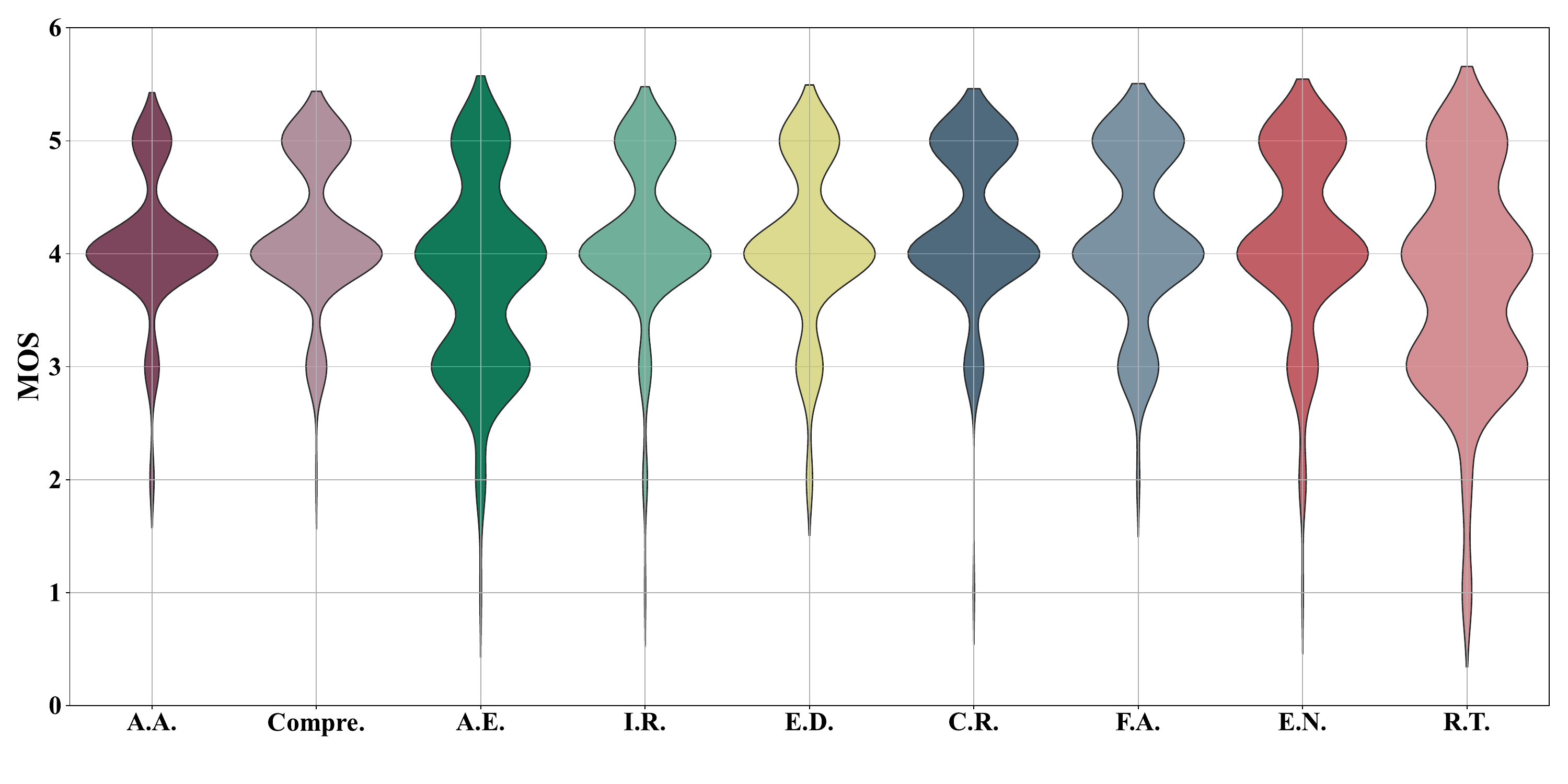} 
    \caption{Deepseek R1} 
    \label{subfig:ds_violin} 
  \end{subfigure}
\begin{subfigure}{0.49\columnwidth}
    \centering
    \includegraphics[width=\linewidth]{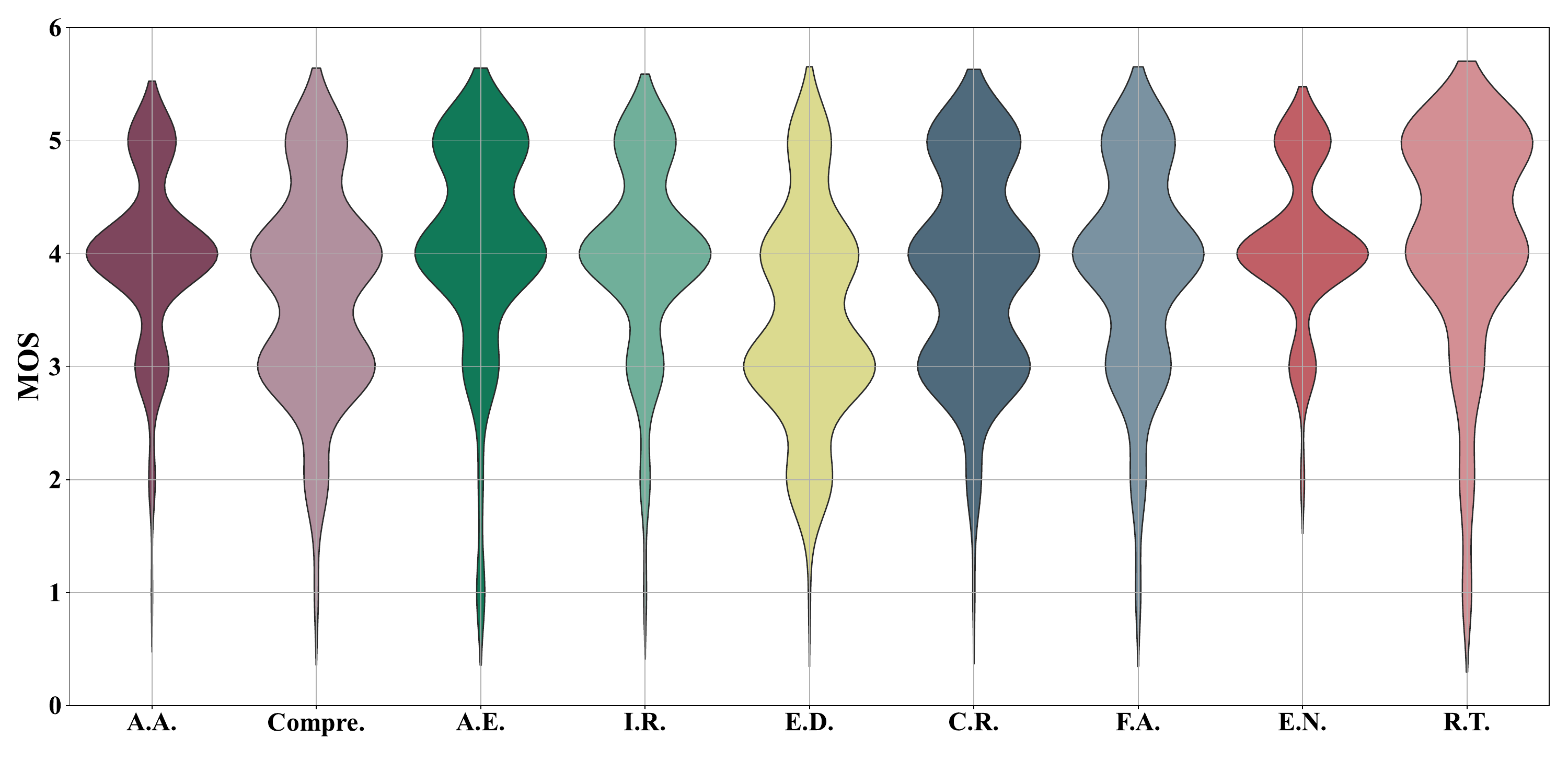} 
    \caption{OpenAI o3 mini} 
    \label{subfig:os_violin} 
  \end{subfigure}
  \begin{subfigure}{0.49\columnwidth}
    \centering
    \includegraphics[width=\linewidth]{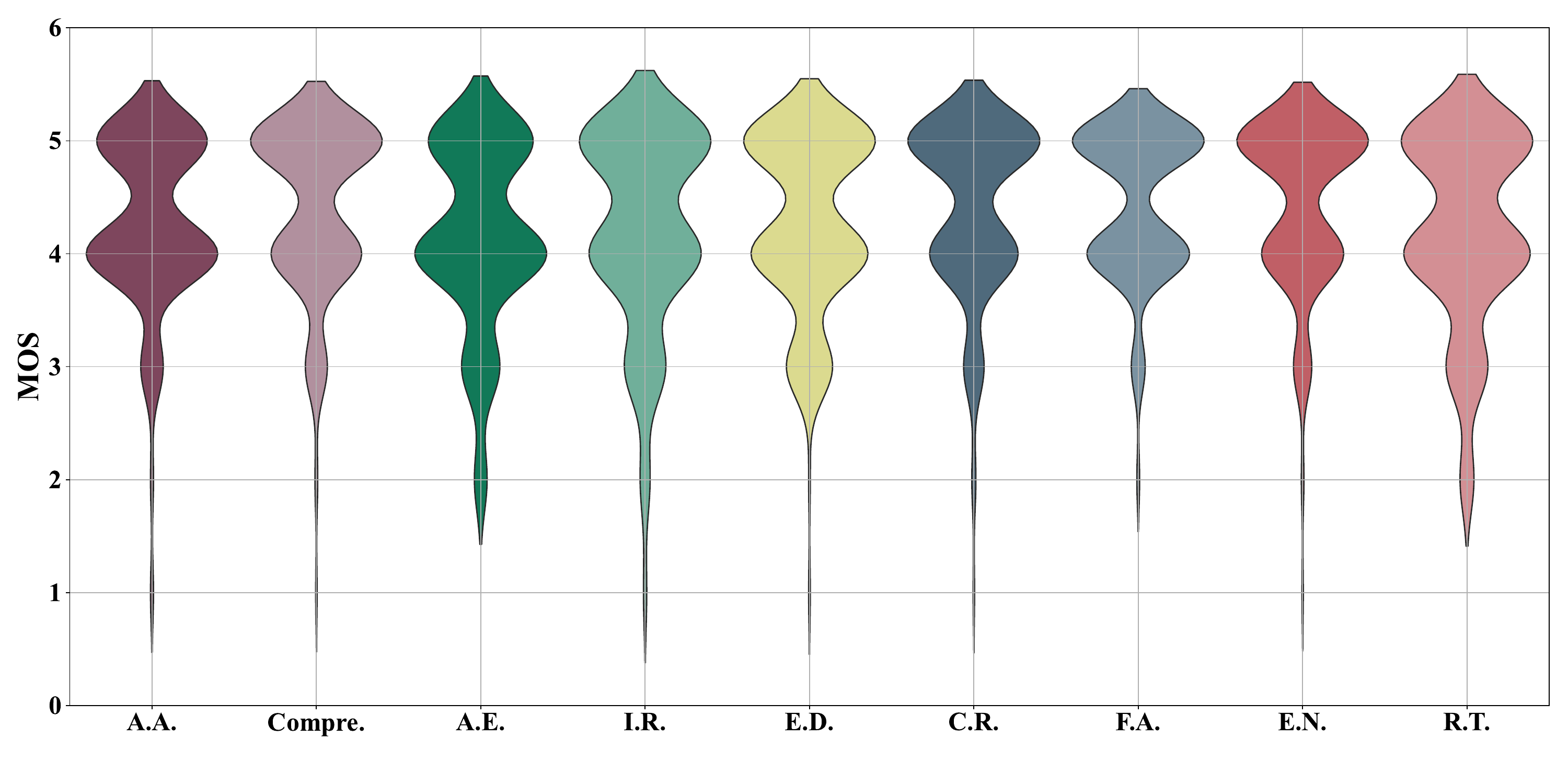} 
    \caption{Grok 3} 
    \label{subfig:grok_violin} 
  \end{subfigure}
\begin{subfigure}{0.49\columnwidth}
    \centering
    \includegraphics[width=\linewidth]{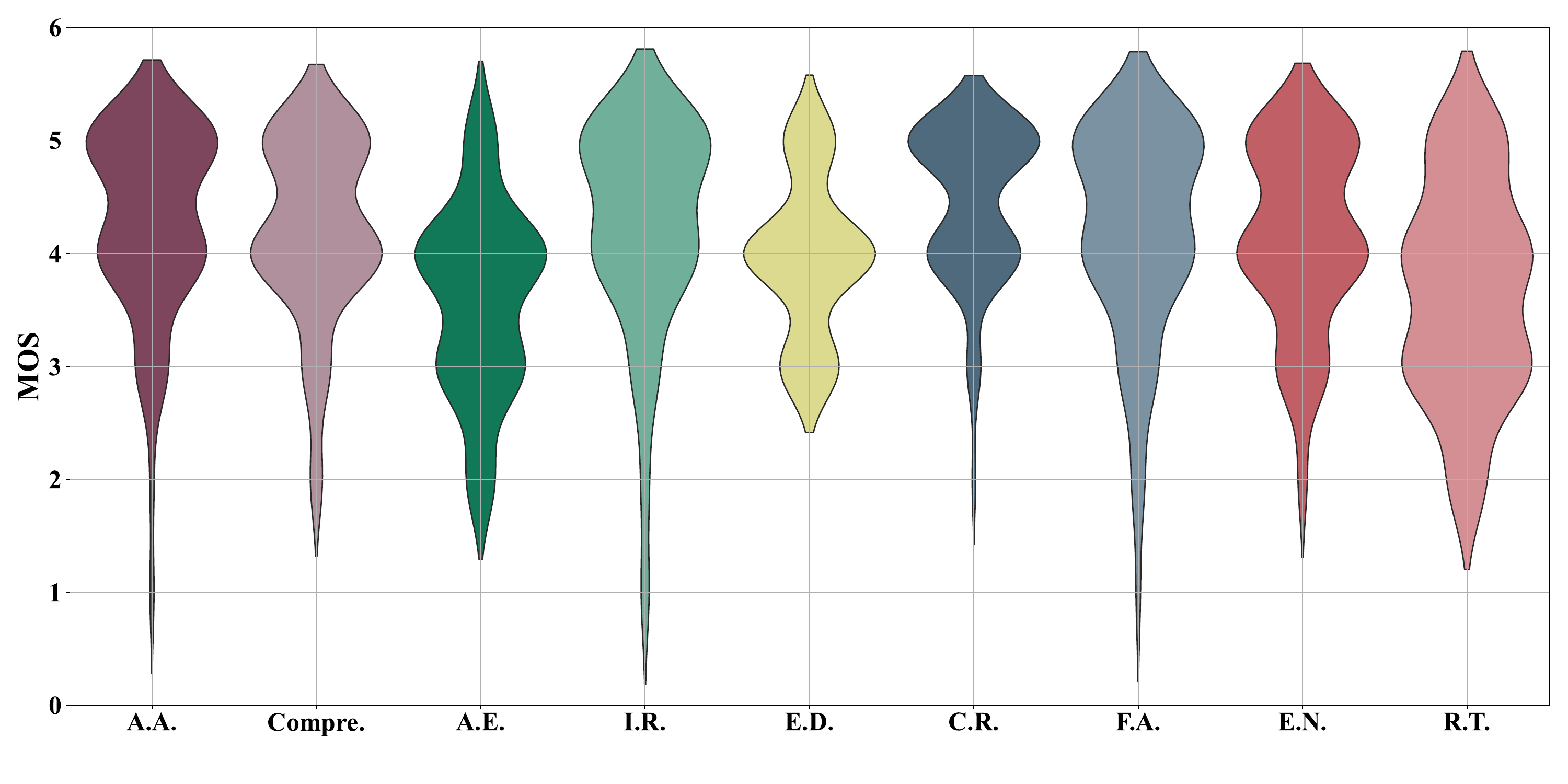} 
    \caption{Gemini 2.5} 
    \label{subfig:gemini_violin} 
  \end{subfigure}
  \caption{Distribution of MOS across different dimensions.}
  \label{fig:violin}
\end{figure}

Here are more detailed results of our subjective experiments. Figure \ref{fig:violin} shows the distribution of MOS across different dimensions for 4 models. Figure \ref{fig:combined results} shows the performance comparison for problem types and languages.



The scoring results in Figure \ref{fig:language} show that while Grok 3 performs consistently well in both Chinese (4.31) and English (4.30), other models like DeepSeek R1 and Gemini 2.5 show slight variations in performance depending on the language. The average model scores are similar in both languages (4.09 in Chinese and 4.10 in English), with no significant difference. Grok 3 maintains the highest scores in both languages, demonstrating strong cross-lingual adaptability.





As shown in Figure \ref{fig:problem}, model performance varies by task type. Grok 3 excels in both literature synthesis and innovation-driven tasks, while DeepSeek R1 performs better on innovation tasks. OpenAI o3 mini shows the largest performance gap, with a 0.20-point advantage on innovation tasks. Gemini 2.5 performs more evenly across tasks, with scores of 4.19 for innovation and 3.98 for synthesis. These results suggest that model choice should align with research goals, with Grok 3 being best for synthesis and other models like DeepSeek R1 and OpenAI o3 mini better suited for creative tasks.

\end{document}